\pdfoutput=1
\documentclass[11pt]{article}
\usepackage[pdftex]{graphicx}
\usepackage[T1]{fontenc}
\usepackage[utf8]{inputenc}
\usepackage{microtype}

\usepackage{acl}
\usepackage{times}
\usepackage{latexsym}
\usepackage{epsfig}
\usepackage{amsmath}
\usepackage{amssymb}
\usepackage{subcaption}
\usepackage{tabularx}
\usepackage{booktabs}
\usepackage[capitalize]{cleveref}

\graphicspath{{img/}}

\usepackage{floatrow}
\setkeys{Gin}{width=\linewidth}

\usepackage{xspace}

\makeatletter
\DeclareRobustCommand\onedot{\futurelet\@let@token\@onedot}
\def\@onedot{\ifx\@let@token.\else.\null\fi\xspace}

\def\eg{e.g\onedot} \def\Eg{E.g\onedot}
\def\ie{i.e\onedot} 
\def\cf{cf\onedot} 
 \def\vs{vs\onedot}

\makeatother

\def\fiber{{\sc FIBer}}

\title{\fiber{}: Fill-in-the-Blanks as a Challenging \\Video Understanding Evaluation Framework}

\author{Santiago Castro \quad Ruoyao Wang \quad Pingxuan Huang \quad Ian Stewart \\
\textbf{\quad Oana Ignat \quad Nan Liu \quad Jonathan C.\ Stroud \quad Rada Mihalcea} \\
University of Michigan -- Ann Arbor, USA \\
\texttt{sacastro@umich.edu}
}

\begin{document}

\twocolumn[{%
\renewcommand\twocolumn[1][]{#1}%
\maketitle
\begin{center}
  \footnotesize
  \captionsetup{type=figure}
  \begin{tabular}{p{0.31\textwidth} p{0.31\textwidth} p{0.31\textwidth}}
    \includegraphics{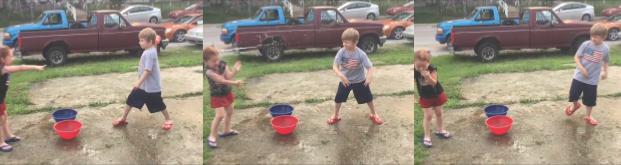} & \includegraphics{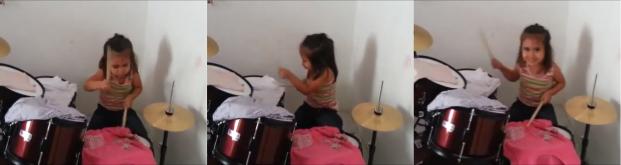} & \includegraphics{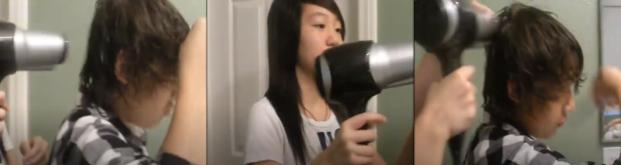} \\
    {Two children throw \_\_\_\_\_ at each other as a video is captured in slow motion.} & {\_\_\_\_\_ sits at a drum set and practices playing the drums.} 
    & {A boy is trying to comb his hair while \_\_\_\_\_ dries it.} \\
    {\textbf{Correct answers}: balloons, balloons filled with water, balloons of water, pink balloon, pink water balloon, things, water, water balloons, water-filled balloons} & {\textbf{Correct answers}: child, drummer, future drummer, girl, kid, little girl, little kid, musician, small child, young girl} & 
    {\textbf{Correct answers}: another person, friend, girl, his sister, his sister with hairdryer, person, young woman} \\
    \end{tabular}
    \captionof{figure}{Three examples from the \fiber{} dataset, each including three video frames, the caption, the blanked answers from the original caption together with the collected answers (all answers normalized, see \cref{sec:data_annotation}).}
\label{fig:dataset-examples}
\end{center}%
}]

\begin{abstract}
We propose fill-in-the-blanks as a video understanding evaluation framework and introduce \fiber{} -- a novel dataset consisting of 28,000 videos and descriptions in support of this evaluation framework. The fill-in-the-blanks setting tests a model's understanding of a video by requiring it to predict a masked noun phrase in the caption of the video, given the video and the surrounding text. The \fiber{} benchmark does not share the weaknesses of the current state-of-the-art language-informed video understanding tasks, namely: (1) video question answering using multiple-choice questions, where models perform relatively well because they exploit linguistic biases in the task formulation, thus making our framework challenging for the current state-of-the-art systems to solve; and (2) video captioning, which relies on an open-ended evaluation framework that is often inaccurate because system answers may be perceived as incorrect if they differ in form from the ground truth.
The \fiber{} dataset and our code are available at \url{https://lit.eecs.umich.edu/fiber/}.
\end{abstract}

\section{Introduction}

Despite current progress on multimodal (textual and visual) representations,
\textit{language-informed video understanding} is still a very challenging task for machine learning systems \cite{Zhang_2021_CVPR,li2021value}. This is due in large part to the task setup and the dataset construction.
Current video understanding datasets often have at least one of two major limitations. First, they have limited application value. \Eg{}, multiple-choice questions~\cite{Lei18,Tapaswi16,Jang17,lifeqa} do not reflect real-world tasks. Second, they are based on subjective evaluation metrics, \eg{}, video captioning ~\cite{tran2016videomcc,Krishna_2017_ICCV,Zhou_Xu_Corso_2018,VATEX}), and are therefore hard to evaluate automatically, as the ground truth can be expressed in different ways.
In this paper, we address these limitations by introducing a new dataset named \fiber{} that collects multiple perspectives on the same video, focusing on noun phrases as a proxy for different entities and their interactions in the video. Our data focuses on recall and tests the ability of models to capture a wide range of possible interpretations for a particular aspect of a video. 

We construct the \fiber{} dataset by systematically blanking captions from an existing video captioning dataset named VaTeX~\cite{VATEX} and by providing additional correct answers for the blanks. VaTeX is a video captioning dataset that contains 40,000 10-second YouTube videos with 10 English captions per video.\footnote{Licensed under Creative Commons, more information here: \url{https://eric-xw.github.io/vatex-website/index.html}.} We build our video fill-in-the-blanks dataset by blanking random noun phrases from one of the English captions for each video, from a subset of VaTeX consisting of 28,000 videos.
Through extensive analyses, we show that the blanked noun phrases are essential for understanding important visual aspects from the video.

To address the fill-in-the-blanks task, we propose a Transformer-based~\cite{vaswani2017attention} multimodal model.
Our experiments show that our best multimodal model achieves a token-level F1 score of 71.4 while the F1 score of crowd workers is 82.5, indicating that this task is challenging for video and text understanding.

The contribution of this work is threefold:
(1) We propose a novel fill-in-the-blanks task as an evaluation framework that addresses the drawbacks associated with previous approaches to video understanding.
In support of this framework, we introduce \fiber{}, which is a novel dataset of 28,000 videos and fill-in-the-blanks captions with multiple correct answers.
(2) We propose several unimodal baselines and two multimodal models for solving this task.
(3) We provide a detailed analysis of the data to measure the diversity and complexity of the answers, and also conduct an error analysis of the models' performance, to gain insights into the blanked captions and videos that are hard for the models to solve.

\section{Related Work}

\textit{Language-informed video understanding} is a complex task that has been extensively addressed in the multimodal (natural language and computer vision) machine learning research through diverse tasks and benchmarks.

\paragraph{Multiple-Choice Video Understanding.}

Multiple-choice benchmarks consist of identifying the only correct answer from a set of distractors, where the set of possible answers varies depending on the input. Video Question Answering (Video QA), a popular format, consists of answering questions based on the video content. Numerous multiple-choice Video Understanding benchmarks have been proposed such as TVQA~\cite{Lei18}, MovieQA~\cite{Tapaswi16}, TGIF-QA~\cite{Jang17} (Repetition Action and State Transition tasks), LifeQA~\cite{lifeqa}, PororoQA~\cite{kim2017deepstory}, MarioQA~\cite{Mun17}, VCQA~\cite{Zhu2017}, VideoMCC~\cite{tran2016videomcc}, and ActivityNet QA~\cite{yu2019activitynet}. However, they provide choices and are thus easier to solve than generating arbitrary text. A further drawback is that the performance without the visual input is generally already high as models are able to exploit biases in the dataset \cite{Agrawal_2018_CVPR} or they count on other modalities that overlap in functionality with the visual one.

\paragraph{Video Captioning.}

Video Captioning consists of generating a piece of text that describes a given video. This task can be carried out using multiple datasets such as ActivityNet Captions~\cite{Krishna_2017_ICCV} (also features Dense-Captioning), YFCC100M~\cite{thomee2016yfcc100m}, \cite{Alayrac16unsupervised}, DiDeMo~\cite{Hendricks_2017_ICCV}, MSR-VTT~\cite{Xu_2016_CVPR}, YouCook2~\cite{Zhou_Xu_Corso_2018}, How2~\cite{Sanabria2018How2AL}, HowTo100M~\cite{miech2019howto100m}, VaTeX~\cite{VATEX}, TGIF~\cite{tgif-cvpr2016}, MovieNet~\cite{Huang2020MovieNetAH}, LSMDC~\cite{Rohrbach2017}, TGIF-QA~\cite{tgif-cvpr2016} (Frame QA task). Due to the diversity of captions provided, Video Captioning benchmarks do not present a high human agreement and are thus hard to evaluate automatically with certainty \cite{aafaq2019video}.

\paragraph{Video Understanding Based on Filling Blanks.}
VideoBERT~\cite{sun2019videobert}, CBT~\cite{sun2019learning}, UniVL~\cite{Luo2020UniViLMAU}, ActBERT~\cite{zhu2020actbert}, and HERO~\cite{li2020hero} methods propose masking random parts of the input from text and video pairs for training. However, they do this only for the purpose of system training and do not use the framework to test and evaluate video understanding. The only exception is MovieFIB~\cite{Maharaj17} which employs a video fill-in-the-blanks scheme, based on LSMDC~\cite{Rohrbach2017} for both training and evaluation. However, these methods have several drawbacks. They blank a single word, which makes it easier to guess; they evaluate correctness with a single ground-truth answer per caption; and %
they focus on the movies domain (we focus on YouTube videos).

\paragraph{Concurrent Work.}

The most similar work to ours is VidQAP~\cite{vidqap}, which presents an evaluation framework to fill in blanks with phrases using semantic roles based on ActivityNet Captions~\cite{Krishna_2017_ICCV} and Charades~\cite{charades}; unlike this existing work, we design our benchmark to feature a high human accuracy (avoiding ActivityNet Captions as it is contextualized, collecting multiple correct answers, and showing a high human performance). Our work is also close to \cite{just_ask} on evaluating the use of free-form QA; however, they employ a small vocabulary and no human accuracy that serves as an upper bound for the task.

The novelty of our work lies in our use of a hard task (a considerable gap between human and best model performance) that measures a form of video understanding while at the same time yielding a high human performance due to the large number of possible correct answers we collected (\(\sim\)13 per caption) from multiple annotators (\(\sim\)9 per caption).

\section{Video Fill-in-the-Blanks Dataset}%
\label{sec:data}

We construct \fiber{} -- a large video understanding dataset that can evaluate the ability of a model to interpret and use a multimodal context by requiring the models to ``fill in'' (generate) a ``blank'' (a missing constituent) in this context. We build \fiber{} by following two main steps: (1) data generation, where we compile a large set of video-caption pairs with selectively blanked words; and (2) data annotation, where crowd workers provide additional valid answers for these blanks.

Note that we could also develop a fill-in-the-blanks dataset by completing only the first step: the data generation. However, this would result in only one valid answer (the original blanked word or phrase), which can lead to unfair evaluations that are too strict because of alternative correct answers being dismissed (\eg{}, ``child'' provided as an answer where the blanked word was ``kid''). Other than manual annotations, we found no high-quality method to automatically obtain additional correct answers. For example, ``building'' and ``t-shirt'' in \cref{tab:qualitative_analysis} are too dissimilar but both are correct, ``pink'' and ``yellow'' in \cref{fig:dataset-examples} are semantically close but only one is correct.

\subsection{Data Generation}

The dataset is constructed starting with the VaTeX~\cite{VATEX} dataset. VaTeX is a multilingual video captioning dataset, consisting of over 41,250 video clips, each of which is taken from a unique public YouTube video, and lasts around 10 seconds. For each video clip, there are 10 English and 10 Chinese captions associated with it.

We produce blanked captions by blanking noun phrases in the English captions in VaTeX. We chose to mask only noun phrases for three main reasons. First, noun phrases often require visual information for identification or understanding. They cover a large variety of information regarding visual content, as their head nouns can describe people, objects, scenes, events, and more. A model often needs to identify the related objects in the videos, as well as the properties of objects (\eg{}, color, number, or size) to fill the blank correctly.

Second, nouns are usually essential to understanding of \textit{visual} content and serve as reliable predictors of the ability of a system to understand a video. Other phrases, such as verbs or adjectives, can more easily be guessed from the text only while ignoring the visual information. To illustrate, consider the example ``A woman \_\_\_\_\_ in the pool,'' where a model can easily predict that the blank should be ``swims'' from the textual content only, which would not be the case for ``A woman swims in \_\_\_\_\_'', where the blank could be completed by sea, pool, lake, water, and other similar nouns.

Third, in preliminary experiments, we found that nouns lead to more robust annotations as compared to e.g., adjectives, which can have low inter-annotator agreement due to their subjectivity. As an example, consider the phrase ``A \_\_\_\_\_ hill stands behind the house.'' where the blank could be filled with a color property, a size property, or another attribute.

For each video, we choose the first English caption that contains at least one noun phrase as detected by spaCy\footnote{We used the model \texttt{en\_core\_web\_trf} from spaCy v3. An error analysis identified only three tagging errors in a sample of 247 sentences.}~\cite{spacy}, and randomly blank one of these noun phrases to generate an instance. Accordingly, we generate our training, validation, and test data starting with the VaTeX v1.1 training set, a random subset of size 1,000 from the validation set, and a random subset of size 1,000 from the test set, respectively.

\subsection{Data Annotation}
\label{sec:data_annotation}

We performed a crowdsourced annotation procedure to collect additional correct answers for each blank in the validation and test sets. As highlighted earlier, the main reason for collecting these additional annotations is to reflect the natural diversity of language, and have multiple alternative answers for each blank.

\begin{figure}
  \centering
  \includegraphics[scale=0.9]{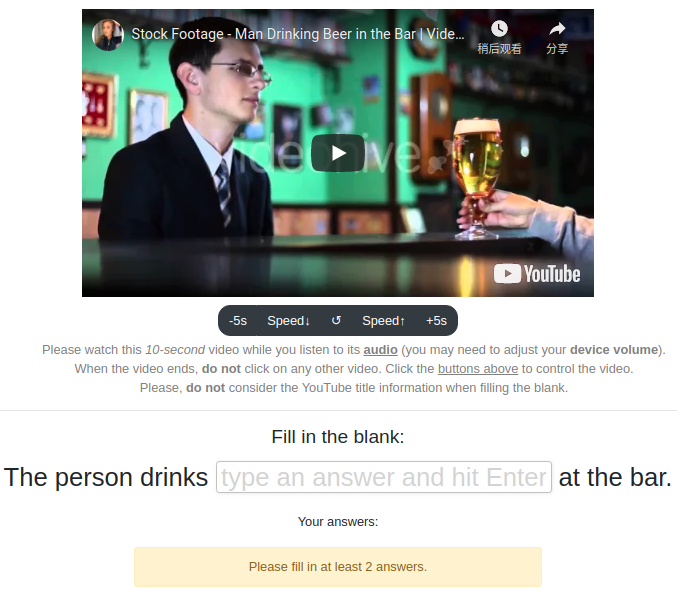}
  \caption{Annotation interface.}
\label{fig:interface}
\end{figure}

We use Amazon Mechanical Turk (AMT) for the annotation. \Cref{fig:interface} shows the annotation interface and a highlight of the data collection instructions (additional guidelines were provided, not shown here for space reasons). For each blanked caption, workers were presented a video clip along with the corresponding masked caption. They were then asked to fill in the blank with a noun phrase.\footnote{We blanked multi-word spans for the task, rather than single-word noun phrases, because blanking a single noun at a time led to a lower annotator agreement in preliminary experiments, likely due to the lower likelihood of overlap. For example, annotator 1 might write ``young boy'' and annotator 2 might write ``young child'', which would have at least some overlap as compared to ``boy'' and ``child'' (no overlap).} We also asked annotators to provide answers in a confidence-descending order (the first answer should be the most natural one to the annotator).

We presented five videos in each Human Intelligence Task (HIT). Nine workers annotated each of them with at least two answers for each blank. We paid a bonus for each extra answer for each blanked caption, from the second one to the fifth one, to encourage them to provide more answers. We calculated a \$12 hourly rate for a worker that provides at least five answers. We estimated the time to annotate one video to be 30 seconds. Consequently, the HIT pay rate was \$0.2, which could result in a total of \$0.5 with the added bonus. Additionally, we offered another type of bonus of \$0.2 to the worker with the largest number of correct answers for every HIT, to encourage them to provide more than five answers.

We required workers to be in Canada or the United States,\footnote{We restricted the task to these countries because it is a good proxy for proficient English speakers and because our task received lower-quality responses otherwise.} and to have completed at least 1,000 HITs on AMT with at least a 92\% approval rate.
The interface also checked that for a given worker and caption the answers were different.
For this, we first normalized the answers by lower-casing, stripping punctuation and extra spaces, and removing the determiners ``the'', ``a'', and ``an.''

During the annotation, we manually reviewed a sample to identify cases of incorrectly tagged noun phrases (\eg{}, ``inside'' marked as a noun when it should be a preposition) and factually incorrect noun phrases (\eg{}, referring to bags as ``eggs'' without any information on the contents of the bags); we disqualified workers who consistently provided incorrect annotations.
After collecting annotations, we filtered for noun phrases using the same method as before, based on whether the text is parsed as a noun phrase (including bare nouns, \eg{} ``\underline{man} is walking''), a wh-phrase (``who is speaking''), a simple gerund (``\underline{eating} is a good way to stay healthy''), or infinitive (``\underline{to eat} is wonderful'').

\begin{table}
    \small
    \begin{tabular}{p{3cm}rr}
        Statistic & Original phrases & Annotated \\
        \toprule
        Noun phrases (before filtering) & 100\% & 95\% \\
        Unique answers per caption & $\sim$ & 13.0 $\pm$ 4.14 \\
        Unique answers per caption per annotator & $\sim$ & 2.63 $\pm$ 0.49 \\
        Characters per token & 5.09 $\pm$ 1.89 & 5.27 $\pm$ 2.00 \\
        Tokens & 1.47 $\pm$ 0.68 & 1.36 $\pm$ 0.68 \\
        Visual word use (color, number, or size) & 8.21\% & 3.31\%
    \end{tabular}
    \caption{Summary statistics for the originally blanked phrases and the annotated answers. The token counts are computed after the text normalization. The statistics for the annotated answers correspond to the ones after filtering for noun phrases (see \cref{sec:data_annotation}), except for the noun phrases percentage.}
\label{tab:label_annotation_statistics}
\end{table}

We compute summary statistics on the annotated data to determine the degree of similarity with the originally blanked phrases.
The statistics are shown in \cref{tab:label_annotation_statistics}.
We find that, in general, annotators tend to provide \(\sim\)3 unique answers for the provided data.
Compared to the original phrases, annotators tend to use about the same number of tokens.
Annotators also use visual words at a much lower rate than the original phrases, possibly because the task encouraged the annotators to generate as many distinct nouns as possible without regard to descriptive information.

\subsection{Data Analysis}
\label{sec:annotation_analysis}

To further validate the utility of the annotations collected in this study, we provide an extensive analysis of the answers (which is obtained from the union of the annotations and the originally blanked phrases).

We compute the most-frequent answers and find, as expected, that noun phrases related to ``person'' are the most frequent: the word ``man'' appears in 5.7\% of total original phrases and 1.2\% of total annotations (see \Cref{fig:top_k_label_annotation_nouns} in the Appendix). Note that our annotations have a long tail distribution, as the most-frequent noun phrase appears in only 1.2\% of total annotations. In addition, we find that answers related to ``person'', such as ``another person'' are not trivial. On the contrary, in the third example in \cref{fig:dataset-examples}, for example, a model has to reason about the actions of both persons and distinguish between them. The other two examples in \cref{fig:dataset-examples} also reflect how a model needs to understand both the video and the text in order to complete the blanks.

\begin{figure}
    \centering
    \includegraphics{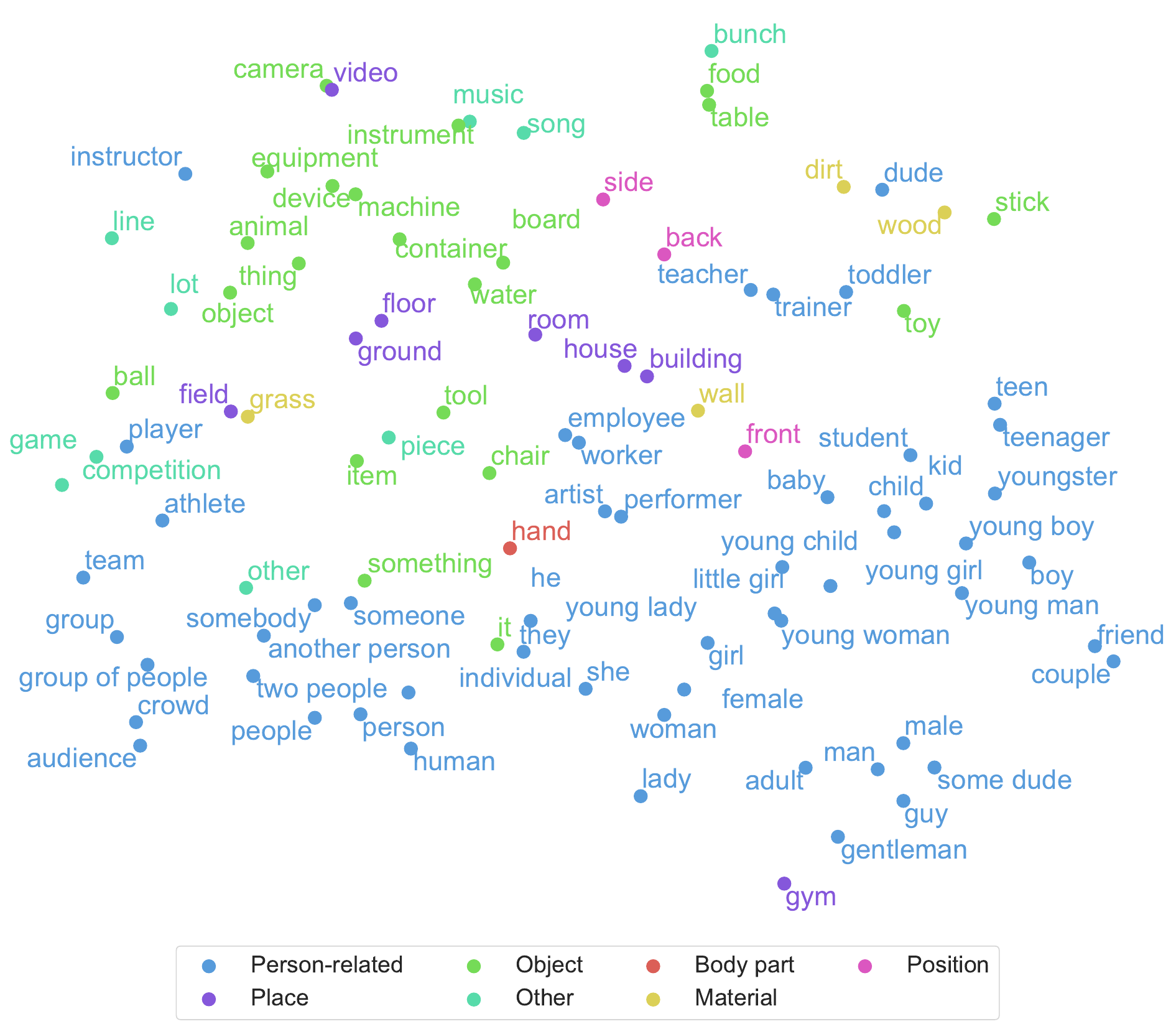}
    \caption{The 2D t-SNE~\cite{Maaten08visualizingdata} representation of the clustering of the top 100 most frequent answers provided for the  blanks. The answers are first converted to singular form, to avoid showing redundant information. The answers are represented using the pre-trained model \texttt{stsb-roberta-base}~\cite{roberta} with Sentence-BERT~\cite{reimers-2019-sentence-bert}. Each color represents a different cluster. The answers are manually mapped to the clusters by one of the authors.}
    \label{fig:cluster}
\end{figure}

\Cref{fig:cluster} shows what kind of answers are depicted in the videos.
This analysis shows the diversity and complexity of answers that a model needs to fill in, demonstrating a strong video understanding. As expected, the cluster \textit{Person-related} has the most answers, followed by the clusters: \textit{Objects} (\eg{}, shoes, glasses), \textit{Places} (\eg{}, mountain, street), \textit{Materials} (\eg{}, metal, wood), and \textit{Body parts} (\eg{}, fingers, head). 
Note also that the \textit{Person-related} cluster, among more typical answers such as ``male'' and ``female'', also contains complex and diverse answers such as ``dancer'', ``workers'', ``musician'' or ``audience''.

\subsection{Human Agreement}
\label{sec:human_agreement}

To establish a reference for the machine models, we compute the agreement among annotators using the evaluation metrics described in \cref{sec:eval_metric}, which we also use for model evaluation (\cref{sec:results}).

Specifically, we  apply a leave-one-out strategy to construct the ``test set'' and the ``ground truth set.'' 
We compare the first answer provided by each crowd worker (which is their most natural/confident answer) against the complete set of answers provided by the other crowd workers, using maximum F1 score (token overlap) and maximum exact match (EM) as agreement metrics, as described in \cref{sec:eval_metric}.

\cref{tab:agreement_stats} shows the inter-annotator agreement. We show the mean values of the agreement metrics per-caption and per-answer (recall there are multiple answers per caption, so in the former case we first average among the answers within the caption and then across the captions).
The higher rates of agreement at the caption level, compared to the answer level, indicate a high amount of answer diversity among the workers.

\begin{table}
    \centering
    \small
    \begin{tabular}{lr} & \\
        Statistic & \% \\
        \midrule
        F1 first answers (per caption) & 82.6 (\(\pm\) 15.7) \\
        Exact Match first answers (per caption) & 75.3 (\(\pm\) 19.7) \\
        F1 first answers (per answer) & 70.0 (\(\pm\) 11.9) \\
        Exact Match first answers (per answer) & 58.1 (\(\pm\) 16.3)
    \end{tabular}
    \caption{Agreement statistics for answers  (leave-one-worker-out-comparison; std. dev. in parentheses).}
\label{tab:agreement_stats}
\end{table}

To validate the quality of the crowdsourced annotations, we also compare them against human annotations collected from two trusted annotators (both researchers at the University of Michigan).
We sample 200 captions from the validation set and ask these two annotators to perform the same labeling task that the MTurk workers performed, and then compare their agreement with the crowdsourced data.
The annotators obtain a per-caption average of 90.2\% F1 score and 49.0\% exact match accuracy, comparable to the agreement scores of the workers.

\subsection{Limitations}

We identify several limitations of our benchmark, which can be the objective of future work.

\paragraph{NPs \vs{} other phrases.} By looking at a video and filling a blank caption with a noun phrase can sometimes indirectly capture other aspects such as actions (verbs, adverbs) and object quality (adjectives, modifiers). However, this is not always the case. This is especially true for noun phrases that are easier to guess (\cf{} \cref{tab:semantic_category_prediction_scores}).

\paragraph{Focus on human actions.} Our data focuses mostly on human-related activities (\eg{}, sports), and may lack general representation available in other datasets related to animals, nature, and technology, to name a few.

\paragraph{Availability of the videos.} As we build upon VaTeX~\cite{VATEX} and YouTube, some videos may become unavailable over time. To mitigate this issue, the VaTeX website offers to download pre-extracted video features.\footnote{\url{https://eric-xw.github.io/vatex-website/download.html}}

\paragraph{Efficiency of the data annotation process.} Not all videos have multiple possible captions for noun phrases. For example, ``the fork'' may be the only reasonable answer for a given video and blanked caption, and annotators may not have anything else to add.

\section{Multimodal Method for Video Fill-in-the-Blanks}

We propose an encoder-decoder multimodal method to perform the task of video fill-in-the-blanks. We first encode the text and visual modalities together to obtain a semantic representation of the blanked caption and video. The decoder uses the semantic representation to generate text corresponding only to the answer to the blank. To correctly generate an answer, a model needs to learn which parts of videos relate to the missing parts of the caption. To accomplish this, we use the original Transformer architecture~\cite{vaswani2017attention}, whose self-attention mechanism is particularly effective for encoding relations within an input sequence and  
have been shown to perform well in many language understanding tasks.

\begin{figure}
  \centering
  \begin{subfigure}{\textwidth}
    \includegraphics[scale=0.2]{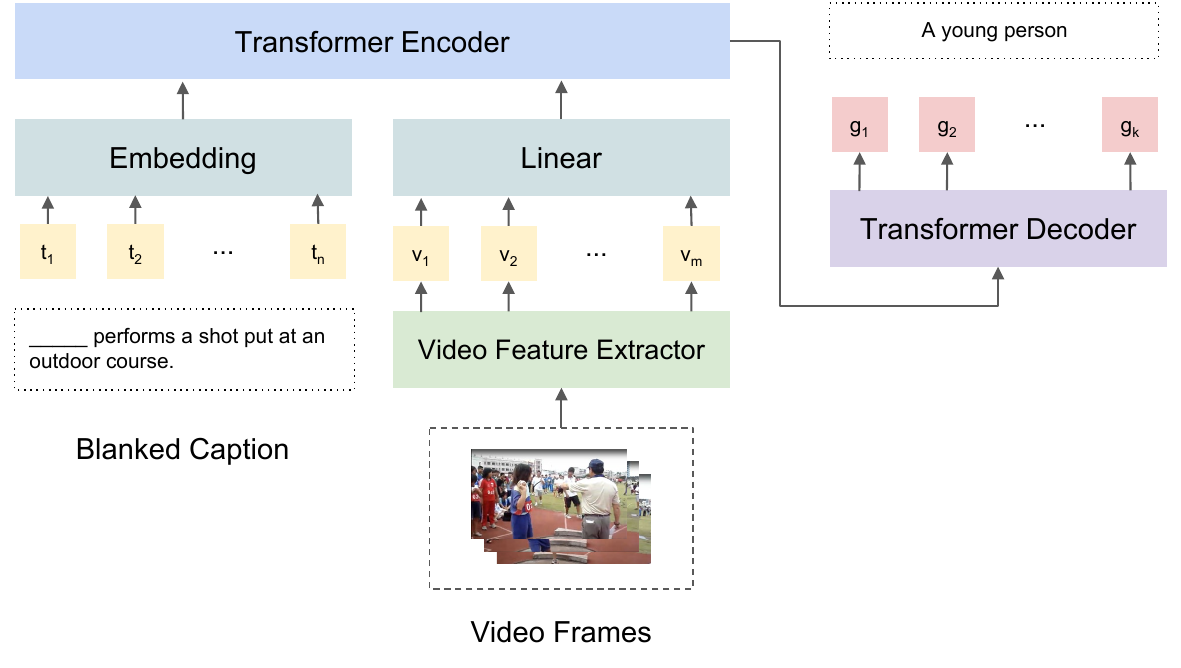}
    \caption{}
    \label{fig:mm_model_t5}
  \end{subfigure}
  \begin{subfigure}{\textwidth}
    \includegraphics[scale=0.2]{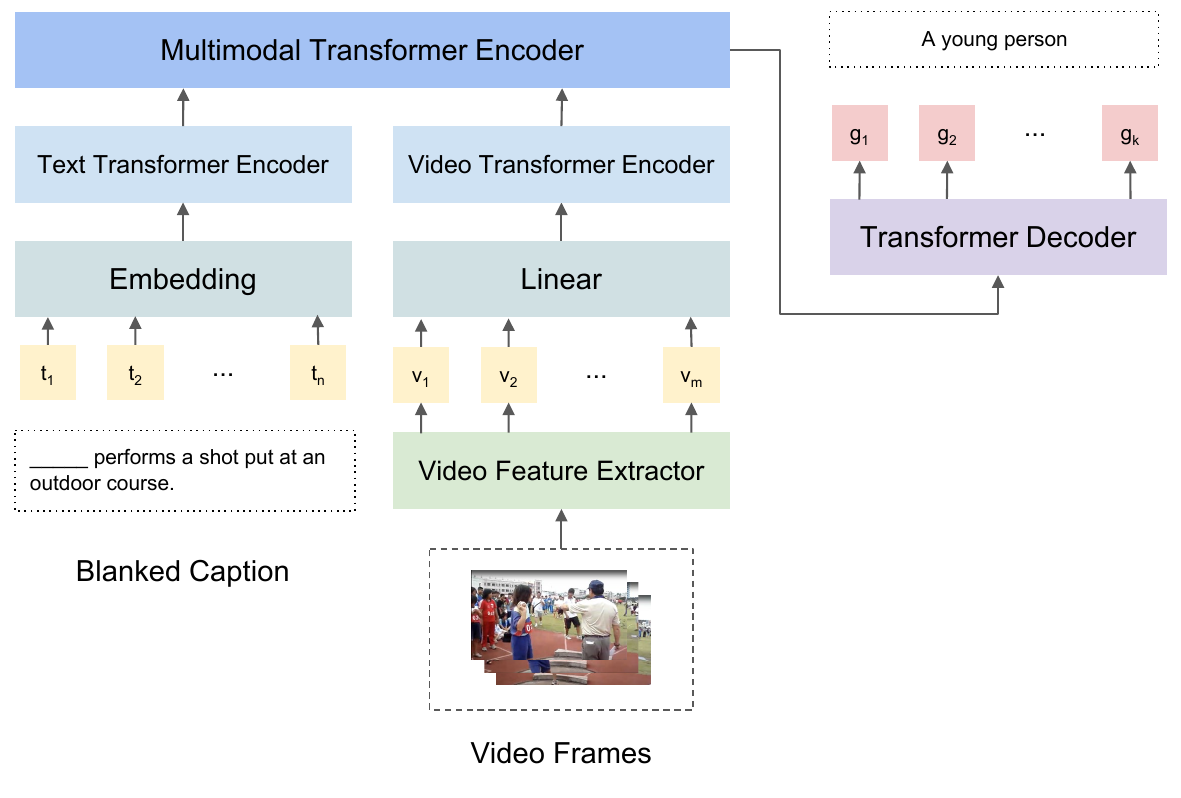}
    \caption{}
    \label{fig:mm_model_t5_two_stream}
  \end{subfigure}
  \caption{(a) Early-fusion multimodal model for video fill-in-the-blanks. (b) Bate-fusion multimodal model for video fill-in-the-blanks.}
  \label{fig:mm_models}
\end{figure}

We consider two types of encoders, namely the early-fusion encoder and the late-fusion (two-stream) encoder. The structure of our multimodal model with an early-fusion encoder is shown in \cref{fig:mm_model_t5}. The input to the model consists of the tokenized blanked caption-text \(t_1, \dots, t_n\), as well as a representation of the video consisting of multiple video sequence features \(v_1, \dots, v_m\) from a video feature extractor. The blanked captions are embedded by an embedding layer.
The video features are projected into the encoder by a linear layer. We use a special token to represent the masked phrase and another one to separate the input text and video sequences. We add positional embeddings to each input token or video feature to represent the sequence order, and another embedding to indicate whether it belongs to the text or video sequence similarly to BERT~\cite{devlin2019bert}.

The late-fusion model is shown in \cref{fig:mm_model_t5_two_stream}. The late-fusion model encodes the language and video first separately and then jointly. This is because the modalities may benefit from learning independently about their own context before using them together.

\subsection{Implementation Details}

For the video encoder, we use the existing I3D~\cite{carreira2017quo} features (size 1024 every 8 consecutive frames) provided by the VaTeX dataset~\cite{VATEX}, in which videos were sampled at 25 fps.
We initialize our multimodal model using T5~\cite{t5}, given its ability to fill in variable-length blanks.
T5 is an encoder-decoder Transformer~\cite{vaswani2017attention} model that is a good starting point as it provides state-of-the-art performance on text-only tasks and it was pretrained to fill arbitrary-length text spans that were previously masked.
Building upon T5 allows our model to not only leverage the pre-trained large-scale language models that already have strong language abilities but also to fuse it with visual inputs. We initialize the early-fusion model with pretrained \texttt{T5-base} weights. For the late-fusion model, we use \texttt{T5-base} for the text encoder and for the decoder. We use two one-layer transformers to encode videos and fuse text and video features, and the weights of these two transformers are randomly initialized. Following T5 model implementation, the special token \texttt{<extra\_id\_0>} is used to represent the blanked phrase, and \texttt{<\textbackslash s>} is used to separate the text and video sequences. The generated output follows T5 output format: the special token \texttt{<extra\_id\_0>} followed by the predicted text for the blanked phrase. See \cref{sec:more_impl_details} for more details.

\subsection{Baselines}

We compare our model to the following baselines.

\paragraph{Most Frequent Answer.}

The baseline makes use of the most frequent answer in the training set (``a man'') as the answer to all the blanked captions during evaluation. 

\paragraph{Text-based Transformer.}

Previous visual question answering datasets found that a text-only model can nearly match the performance of the multimodal system~\cite{VQA}.
We analyze the degree to which language alone can contribute to our video understanding framework by conducting experiments based on text-only models. 
We use the off-the-shelf T5-base transformer model~\cite{t5} as our baseline model.
We use both a zero-shot model (not trained on our data) and a fine-tuned model. For the latter, we use the \texttt{base} model v1.1 because it performed better in our experiments on the validation set. 
The decoding hyperparameters are the same as in the multimodal models, except the beam size is 8 for both the zero-shot one and 2 for the fine-tuned variant as we obtained the best validation results for each one using these beam sizes.

\paragraph{Single video feature.}

We consider using a single I3D feature per video to determine how well the model does with a small portion of the video. Based on a study of 50 randomly sampled videos, the blanked entity in the caption appeared 95\% of the time in the third second of the video (see \cref{fig:when} in the Appendix). For this method, we pick the I3D feature which corresponds roughly to it and apply it to the proposed multimodal methods instead of using all the video features. Note I3D takes a window of 16 frames as input, which in our case corresponds to 640 milliseconds, centered at the mentioned moment within the video. This can be seen as a small generalization of the Image Understanding task, which considers a single image (frame).

\section{Experiments and Results}

We perform experiments and evaluations using the dataset described in \cref{sec:data}. 

\subsection{Evaluation Metrics}%
\label{sec:eval_metric}

We use exact match accuracy and ROUGE-1 F1 score (token-level) \cite{rouge} to evaluate the output of the generation models and to evaluate human agreement (\cref{sec:human_agreement}).
For the exact match, we count a generated text string as correct if it has at least one string-level match among the provided annotations.
For the token-level F1, we compute the token overlap (true positives) between the generated text string and each annotation, normalized by the sum of the true positives and average of the false negatives/positives. We then compute the maximum across all annotations.
For all evaluations, we computed the metrics based on the normalized text (\ie{}, without articles).

\subsection{Results}%
\label{sec:results}

We evaluate the visual understanding ability of our multimodal model by comparing its performance with the text-only baseline and the human performance.
The results from the fill-in-the-blanks task are shown in \cref{tab:main-experiments}.
The accuracy of the text-only model and F1 score are low, indicating that the language bias is controlled in our dataset. The multimodal model outperforms the text-only baselines in both exact match accuracy and F1 score, meaning that our multimodal model is able to learn video features relevant to caption language during training.
We also note that the early-fusion multimodal model (T5 + I3D) slightly outperforms the late-fusion multimodal model, which suggests that the model learns more effectively without extra encoders (see \cref{fig:mm_model_t5_two_stream}). Both the early-fusion and the late-fusion multimodal models perform worse with a single I3D feature. This suggests that the model benefits from the whole video to correctly answer the caption.

\begin{table}
    \centering
    \small
    \begin{tabular}{crrrr}
        & \multicolumn{2}{c}{val} & \multicolumn{2}{c}{test} \\
        Method & EM & F1 & EM & F1 \\
        \midrule
        \multicolumn{5}{c}{\sc Baselines} \\
        \midrule
        Most Frequent Answer & 15.4 & 45.1 & 16.4 & 45.3  \\
        \midrule
        T5 zero-shot & 39.3 & 52.0 & 37.4 & 49.2 \\
        T5 fine-tuned & 58.0 & 73.8 & 54.5 & 70.9 \\
        \midrule
        \multicolumn{5}{c}{\sc Our multimodal models} \\
        \midrule
        T5 + 1f I3D & 59.2 & 74.7 & 54.3 & 70.5 \\
        T5 + I3D & \textbf{60.2} & \textbf{75.0} & \textbf{56.2} & \textbf{71.4} \\
        Late-fusion T5 + 1f I3D & 53.7 & 70.3 & 50.3 & 67.6\\
        Late-fusion T5 + I3D & 53.5 & 69.7 & 51.6 & 67.8 \\
        \midrule
        \multicolumn{5}{c}{\sc Upper bound (Human Agreement)} \\
        \midrule
        leave one worker out & 75.3 & 82.6 & 75.0 & 82.5 \\
        new humans* & 49.0 & 90.2 & n/a & n/a \\
    \end{tabular}
    \caption{Results on the validation set. EM stands for Exact Match, and F1 is the token-level F1 score (both in percentage). \emph{1f} refers to the variant of the multimodal model with a single I3D feature. The new humans' performance is measured from a random sample of size 200. See \cref{sec:human_agreement} for more details on the human baselines.}
\label{tab:main-experiments}
\end{table}

We also find a large performance gap between the multimodal model performance and the human performance. Therefore, plenty of space exists for improvements to achieve human performance, and the video fill-in-the-blanks task is worth investigating in future visual understanding research.

\subsection{Error Analysis}

\paragraph{Results per Semantic Label.}
To measure how well the model understands different patterns in the caption data, we compare the predictions generated for blanks corresponding to words of different semantic categories (the rest of the answers generally belong to the same category as the blanked words).
Two of the authors annotated the originally blanked phrases for common non-overlapping semantic categories, including people, passive entities, and locations.

We list the categories and their distribution/size in \cref{tab:semantic_category_prediction_scores}, and we also show the performance for the best text-only zero-shot method (T5 zero-shot), text-only fine-tuned method (T5 fine-tuned), and multimodal method (T5 + I3D). The results of T5 zero-shot show some categories can be easily predicted, without fine-tuning on the dataset, namely \textit{Preposition}, \textit{Pronoun}, and \textit{Event}. However, fine-tuning T5 on our dataset yields improvements for nearly all categories. The multimodal (T5 + I3D) model improves the categories of \textit{Person} and \textit{Abstract} nouns but performs worse for others, namely \textit{Audio} and \textit{Action}. This finding follows from the fact that understanding higher-order audio and visual concepts requires complex reasoning, for which the video-aware model may need more training. In general, \textit{Action} and \textit{Passive entity} will likely require extra attention in future work, considering the comparatively low performance for these categories.

\begin{table}
    \small
    \begin{tabular}{lr|rrr}
        Category & Size (\%) & T5 zs & T5 ft & T5 + I3D \\
        \toprule
        Passive entity & 40.4 & 52.9 & \textbf{63.6} & \textbf{63.6} \\
        Person & 33.4 & 37.0 & 81.8 & \textbf{83.2} \\
        Pronoun & 6.1 & 73.5 & \textbf{85.6} & 84.3 \\
        Location & 5.5 & 55.1 & 74.5 & \textbf{75.4} \\
        Preposition & 4.5 & 81.6 & 95.7 & \textbf{97.5} \\
        Action & 3.9 & 47.8 & \textbf{65.5} & 59.9 \\
        Audio & 2.5 & 56.4 & \textbf{73.0} & 63.6 \\
        Abstract & 2.2 & 59.6 & 70.0 & \textbf{77.9} \\
        Other & 1.5 & 56.9 & 75.0 & \textbf{83.7} \\
        Event & 1.0 & 70.0 & 68.0 & \textbf{84.0} \\
    \end{tabular}
    \caption{F1 scores on the validation set for blanks with different semantic categories, in descending order based on their size. The results correspond to the best T5 zero-shot, T5 fine-tuned, and T5 + I3D models. \textit{Person} corresponds to answers related to people, \textit{Passive entity} represents passive entities such as objects, \textit{Pronoun} includes subject or object pronouns, \textit{Location} corresponds to places in general, \textit{Preposition} includes noun phrases inside prepositional phrases (\eg{}, ``order'' in ``in order to''), \textit{Action} involves activities (``a handstand'' in ``perform a handstand''), \textit{Audio} refers to noun phrases indicated through audio (``the procedure'' in ``the person describes the procedure'', which can only be understood through access to the audio modality), \textit{Abstract} corresponds to high-level concepts (\eg{}, ``a great time''), \textit{Event} are long-running processes (``a party''), and \textit{Other} correspond to instances hard to label for the annotators (\eg{}, ``a video'').}
\label{tab:semantic_category_prediction_scores}
\end{table}

\paragraph{Best Model \vs{} Human Performance.}
To gain insights on how to improve our models for future work, we measure where our best model (T5 + I3D) fails and humans perform well.
We find three main types of wrong predictions. The most common error is predicting ``man'' instead of ``women'', followed by predicting ``person'' instead of ``child'' or ``baby''. The majority of the remaining errors are predictions close to the ground truth answers such as ``dance'' instead of ``exercise'', ``pillow'' instead of ``sheets'', ``rug'' instead of ``sand'', ``floor'' instead of ``court'', ``knife'' instead of ``spatula'' or ``basketball game'' instead of ``wrestling''.

Based on these types of errors, in future work, the model would benefit from pre-training on unbiased data (both gender and age) and also from pre-training on a large-scale multimodal (language and video) dataset, to learn about more diverse situations and objects.

\section{Conclusions}

This paper introduced the fill-in-the-blanks evaluation framework for video understanding. The framework addresses drawbacks of alternative video understanding tasks, such as multiple-choice visual question answering or video captioning.

Our paper makes three important contributions. First, we introduced \fiber{}, which is a large dataset consisting of 28,000 videos and tests based on filling blanks, building upon an existing video captioning dataset with a new set of manual annotations, and using a modified annotation framework to encourage diverse responses among annotators.
This process can be easily replicated to create new fill-in-the-blanks data for other datasets and tasks.
Second, we conducted extensive analyses on the dataset to evaluate the quality of the annotations and to understand the patterns and limitations of the data.
Finally, we introduced a multimodal model that fuses language and visual information and found that the video-aware models significantly outperform the text-only models.
Notably, we found a consistent gap between model performance and human performance, which suggests room for improvement in future models addressing video understanding through the lens of the fill-in-the-blanks task.

The \fiber{} dataset and our code are available at \url{https://lit.eecs.umich.edu/fiber/}.

\section{Ethical Considerations and Broader Impact}

Even though we compensated the annotators based on the quality of the answers they produced (and stated so in the instructions), they were rewarded based on the number of answers they input since we looked for diversity. These incentives may have encouraged the annotators to make many judgments quickly and therefore make biased decisions. Due to these biases, we cannot guarantee that annotators' guesses always match reality. Based on spot-checking, it seems that annotators made reasonable judgments, but others may disagree. We have also observed our data is skewed toward more male noun phrases (\cf{} \cref{sec:gender_representation}), which could be due to a bias both in VaTeX and in the annotators we hired.

Our evaluation weights all errors equally, even though some errors may have a bigger impact than others. For example, someone in a video may be misgendered by being referred to as a ``man'' when the correct reference should be ``woman.''

\section*{Acknowledgments}

We thank Laura Biester for helping with data quality assurance. We thank the following people for reviewing drafts of this document: Artem Abzaliev, Christine Feak, Victoria Florence, Zhijing Jin, and Max Krogius. We also want to thank the \href{https://lit.eecs.umich.edu/}{LIT Research Group @ UMich} members for feedback on some of the ideas discussed here.
This material is based in part upon work supported by the Automotive Research Center (``ARC''). Any opinions, findings, conclusions, or recommendations expressed in this material are those of the authors and do not necessarily reflect the views of ARC or any other related entity.

\bibliography{main}
\bibliographystyle{acl_natbib}

\appendix
\label{sec:appendix}

\section{Dataset}

\subsection{Most-Frequent Noun Phrases}

\begin{figure}
    \includegraphics{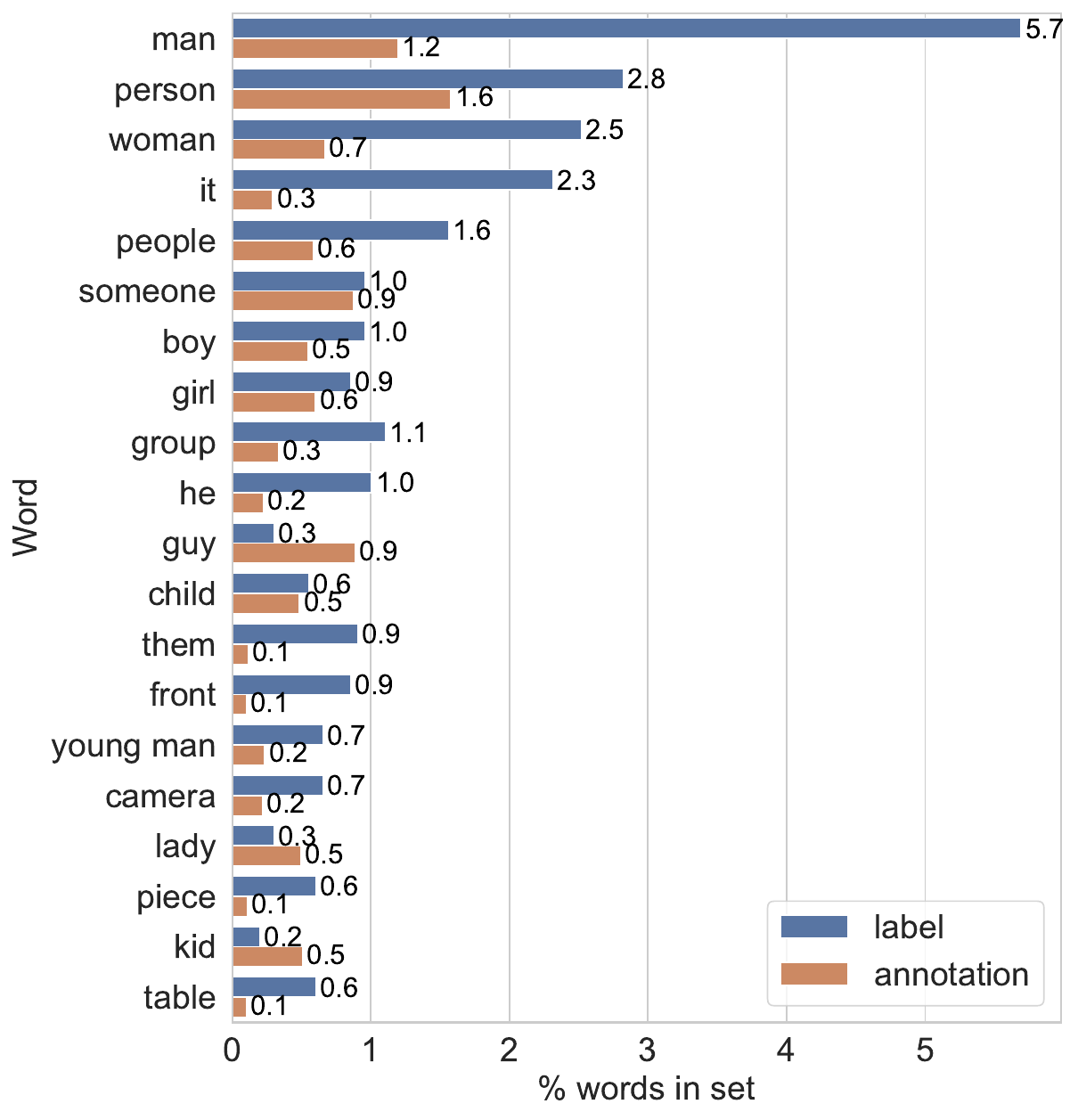}
    \caption{Top 20 nouns for the originally blanked phrases and the annotations in the validation and test data.}
    \label{fig:top_k_label_annotation_nouns}
\end{figure}

We report the most-frequent noun phrases in the original labels and in the annotations we collected, in \cref{fig:top_k_label_annotation_nouns}.
The most frequent nouns for both answer sets tend to reference people, which makes sense considering the content of the videos.
In the annotation data, we see a greater variety of synonyms for the same kind of person (``male'', ``man'', ``guy''), likely a result of the task definition, which encourages paraphrasing.

\subsection{Part-of-speech Distribution}

\begin{figure}
    \centering
    \includegraphics{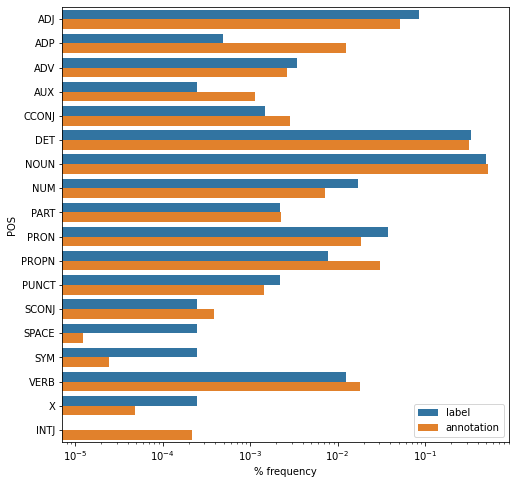}
    \caption{Relative frequency of part-of-speech tags in the originally blanked phrases and the annotated answers.}
    \label{fig:POS_tag_distribution}
\end{figure}

We compare the rate of use of words in different part-of-speech categories for the originally blanked phrases and the annotations, using the same parser specified earlier to label part-of-speech tags in the noun phrases.
The distributions are shown in \cref{fig:POS_tag_distribution}, and we see that the annotations have roughly the same rate of part-of-speech tag use in all categories, except among adjectives and pronouns where the originally blanked phrases have a higher rate of use.
This is likely an artifact of the data collection strategy, which encouraged annotators to generate unique noun phrases rather than phrases with adjectives or pronoun references.

\subsection{Part-of-speech Sequence Distribution}

\begin{figure}
    \centering
    \includegraphics{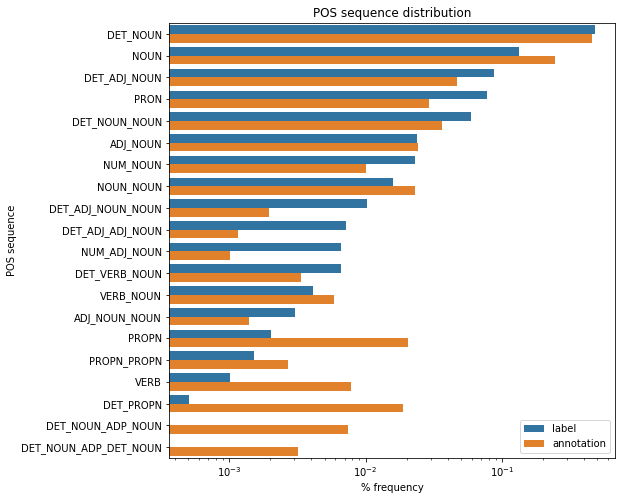}
    \caption{Relative frequency of POS tag sequences in the originally blanked phrases and the annotated answers.}
    \label{fig:POS_tag_sequence_distribution}
\end{figure}

Although the candidate answers collected from crowd workers consist of noun phrases, they may include different part-of-speech (POS) sequences within the noun phrases.
The distributions of POS sequences in \cref{fig:POS_tag_sequence_distribution} show that the annotators tended to write ``bare'' nouns without extra determiners and proper nouns, more than the original phrases.
This makes sense considering that the task asked annotators to provide many unique nouns without consideration for the nouns' structure.

\subsection{Dependency Categories}

\begin{figure}
    \centering
    \includegraphics{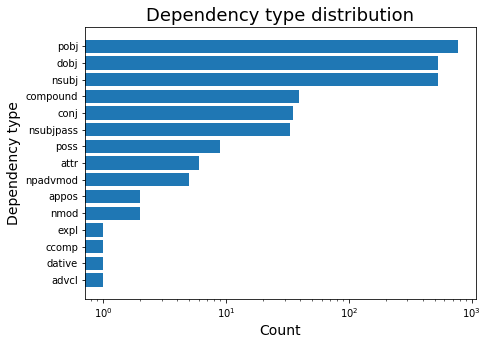}
    \caption{Dependency category counts (per caption).}
    \label{fig:dependency_category_distribution}
\end{figure}

Due to the sampling process, some of the answers occur in different syntactic contexts, \eg{} in a prepositional phrase in ``A woman does push-ups on \_\_\_\_\_'' or as a subject in ``\_\_\_\_\_ at a driving range demonstrating...'' (see \cref{fig:dataset-examples}).
We plot the distribution of dependency categories in \cref{fig:dependency_category_distribution}, which shows that nouns occur in a wide range of positions but mostly occur in a preposition, subject, and direct object positions.

\begin{figure}
    \centering
    \includegraphics{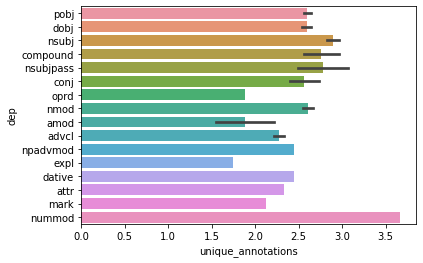}
    \caption{Average number of unique answers per caption, grouped by the dependency category of the root word of the originally blanked phrases. The categories are sorted by their frequency.}
    \label{fig:labels_per_dependency_category}
\end{figure}

Next, we test whether certain syntactic contexts tend to attract more answers from the annotators than others, by computing the mean unique number of answers per annotator within each syntactic context (based on the dependency parse connected to the masked NP).
The distribution is shown in \cref{fig:labels_per_dependency_category}.
Captions that mask noun phrases which occur in preposition (\texttt{pobj}) and direct object (\texttt{dobj}) positions tend to attract slightly fewer unique answers per annotator than the next most-frequent categories, subject (\texttt{nsubj}) and compounds (\texttt{compound}).
This intuitively makes sense, since annotators would likely have fewer options for noun phrases when faced with a preposition or a direct object, as opposed to the less restrictive subject noun position.

\subsection{Gender Representation}
\label{sec:gender_representation}

Often, language processing models can learn to encode social bias due to non-representative training data, such as image captions for photos of men and women taken in stereotypical environments~\cite{zhao2017men}.
We find a slight gender gap in our own data: by using a gender word list, we find that about 10.9\% of the originally blanked phrases are male-related words in contrast to 6.2\% that are female-related, and 9.1\% of the annotations are male-related while 5.9\% are female-related.
We note that the gender imbalance is less severe for the annotations than for the original phrases, and the annotations do in fact use more gender-neutral human words than the labels (6.6\% for annotations vs. 6.0\% for original phrases).
While some of the annotators may undoubtedly have some bias in terms of their decisions, some of the bias may also result from the original video clips.
We acknowledge this limitation as a direction for future work in collecting video caption data.

We used the following lists for gendered words, which were chosen to be in similar semantic categories (\eg{} male ``brother'', female ``sister'', neutral ``sibling''):

\begin{itemize}
    \item Male-oriented words: ``boy'', ``brother'', ``father'', ``guy'', ``he'', ``him'', ``himself'', ``his'', ``male'', ``man'', ``son''
    \item Female-oriented words: ``daughter'', ``female'', ``girl'', ``her'', ``herself'', ``lady'', ``mother'', ``she'', ``sister'', ``woman''
    \item Gender-neutral words: ``adult'', ``baby'', ``child'', ``human'', ``kid'', ``parent'', ``people'', ``person'', ``sibling''
\end{itemize}

\subsection{Spatiotemporal Trends of the Blanked Entities}

\begin{figure}
    \centering
    \includegraphics{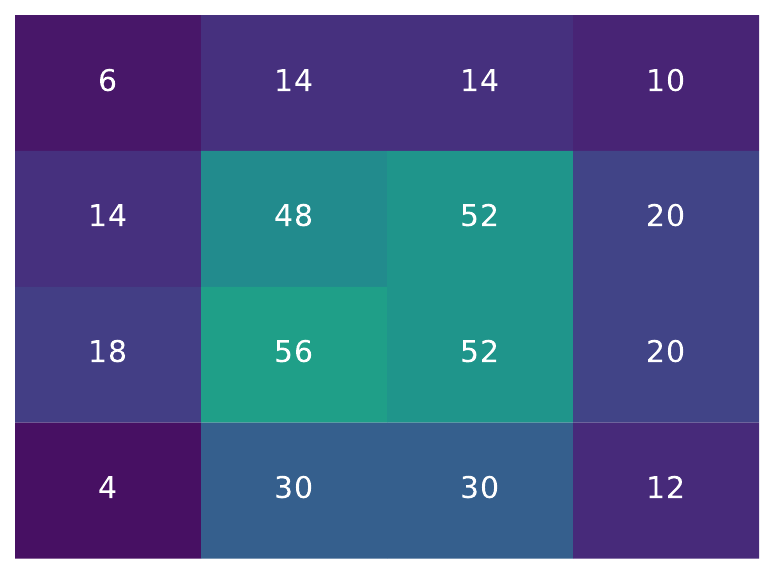}
    \caption{Heat map showing how frequently (\%) the blanked entity appears within a given location of the video, for a sample of 50 videos. Each frame is divided into a 4 by 4 grid. For a given cell, a blanked entity is counted if it touches the cell at any moment of a given video. Note that multiple cells can be counted for a given video because the entity is big enough, or because the entity or the camera moves.}
    \label{fig:where}
\end{figure}

\begin{figure}
    \centering
    \includegraphics{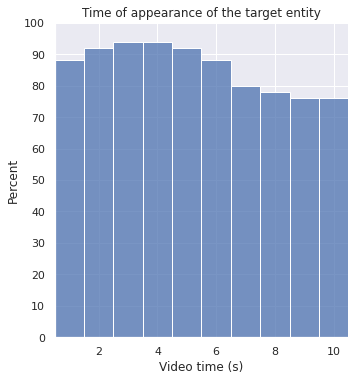}
    \caption{Frequency (\%) that the blanked entity appears at each one-second interval in a given video, for a sample of 50 videos. A time interval is counted if the entity appears at any moment of the one-second duration interval.}
    \label{fig:when}
\end{figure}

\begin{figure}
    \centering
    \includegraphics{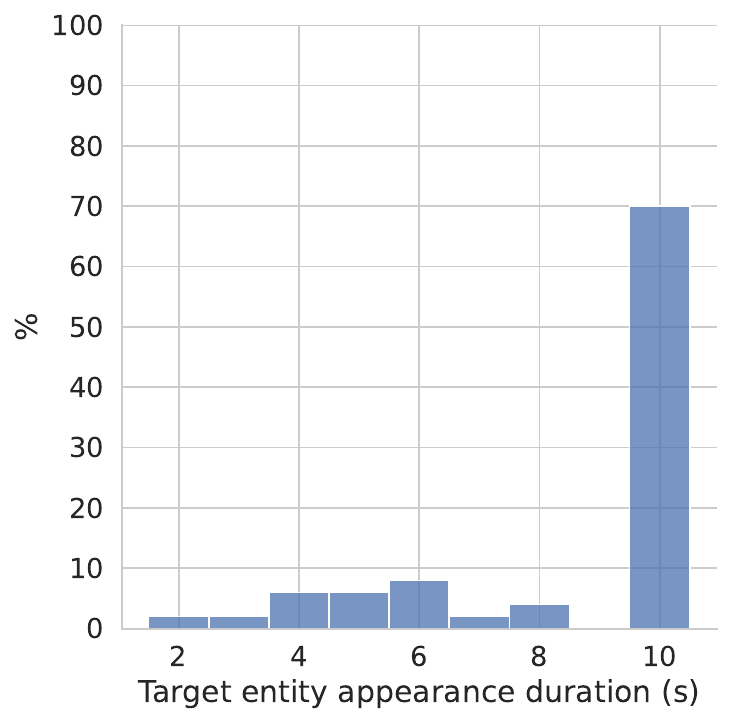}
    \caption{Distribution of the total time that each blanked entity is seen within its video, for a sample of 50 videos.}
    \label{fig:duration}
\end{figure}

One of the authors of this paper randomly sampled 50 videos to analyze spatiotemporal information on the blanked entities. \Cref{fig:where,fig:when,fig:duration} show trends on where, when, and for how long the blanked entities appear in the videos. As expected, the blanked entity generally appears at the center of frames, with a small tendency to be on the lower side. We observe that around 93\% of the time the blanked entity appears between seconds 2 and 4 of the video but that there is still a high chance (75\%) of seeing it at any given moment. 68\% of the time the blanked entities appear for the entire duration of their corresponding video.

\section{Experiments and Results}

\subsection{More Implementation Details}%
\label{sec:more_impl_details}

We use the T5 model from the HuggingFace Transformers library~\cite{wolf2020transformers}.
We train the model with Adam~\cite{Kingma2015AdamAM} on a V100-16Gb with a batch size of 64 for 10 epochs (4,000 steps) using a learning rate of 1e-4 with a warm-up of one epoch and a linear decay. The training time is short, less than an hour. We compute the loss as the cross-entropy between the model-generated output and the originally blanked phrase.

For test-time decoding, we use beam search with a beam size of 4 for the early-fusion model and 8 for the late-fusion one, with a maximum token length of 10. We stop the decoding early, if an example has seen as many complete hypotheses as the beam size (beam search early-stopping\footnote{\url{https://huggingface.co/transformers/internal/generation_utils.html\#transformers.BeamSearchScorer}}). We penalize the repetitions of bigrams within a decoded text. For each example, we choose the first beam that is a noun phrase, as detected by spaCy~\cite{spacy}, or the first one if none. We show the effect of varying the beam size in \cref{sec:beam_search}. We find that modifying the beam search early-stopping property does not lead to major performance changes.

\subsection{Beam Search}
\label{sec:beam_search}

\Cref{tab:beam_search} shows the effect of varying the beam size during the beam search decoding. In all cases, using a beam search of at least size 2 is better than a greedy search. However, the results are marginally better or inconclusive when using beam size 4 or 8. This is probably related to the phenomenon described by Meister \textit{et al.}~\cite{meister2020if} in which beam search does get us closer to the true maximum a posteriori solution but the answers actually start to get worse after a certain point.

\begin{table}
    \small
    \begin{tabular}{l|rrrr}
    & 1 & 2 & 4 & 8 \\
    \midrule
    T5 fine-tuned & 72.9 & \textbf{74.2} & 73.8 & 73.8 \\
    T5 + I3D & 73.0 & 74.0 & \textbf{74.3} & 74.2 \\
    Late-fusion T5 + I3D & 69.0 & 69.6 & \textbf{69.7} & \textbf{69.7} \\
    \end{tabular}
\caption{F1 scores on the validation set for the beam sizes 1 (greedy search), 2, 4, and 8.}
\label{tab:beam_search}
\end{table}

\subsection{Model Size}

In \cref{tab:model_sizes} we show the result of changing the T5 model size for the text-only zero-shot baseline. We note we could not fit the model variant \texttt{t5-11b} into GPU memory. As expected, we note an increase in the evaluation metrics as the model capacity increases.

\begin{table}
    \small
    \begin{tabular}{lrr}
    & EM & F1 \\
    \midrule
    \texttt{t5-small} & 20.2 & 37.1 \\
    \texttt{t5-base} & 34.9 & 50.2 \\
    \texttt{t5-large} & 43.5 & 59.5 \\
    \texttt{t5-3b} & \textbf{44.9} & \textbf{62.6} \\
    \end{tabular}
\caption{Results on the validation set for different model sizes of the T5 text-only zero-shot model.}
\label{tab:model_sizes}
\end{table}

\subsection{Qualitative Analysis}

We show in \cref{tab:qualitative_analysis} several examples of answers correctly predicted by the best multimodal method but incorrectly answered by the best text-only method. Even though the answers provided by the text-only method are plausible by just looking at the text, they do not make sense with the given videos. In the second example, one can quickly tell the person is not at a gym but instead is in some kind of indoor room. For these examples, the multimodal method seems to have identified what is visually important.

\begin{table*}
    \footnotesize
    \begin{tabular}{p{1.5cm}|p{4.1cm}p{4.1cm}p{4.1cm}}
        & \includegraphics{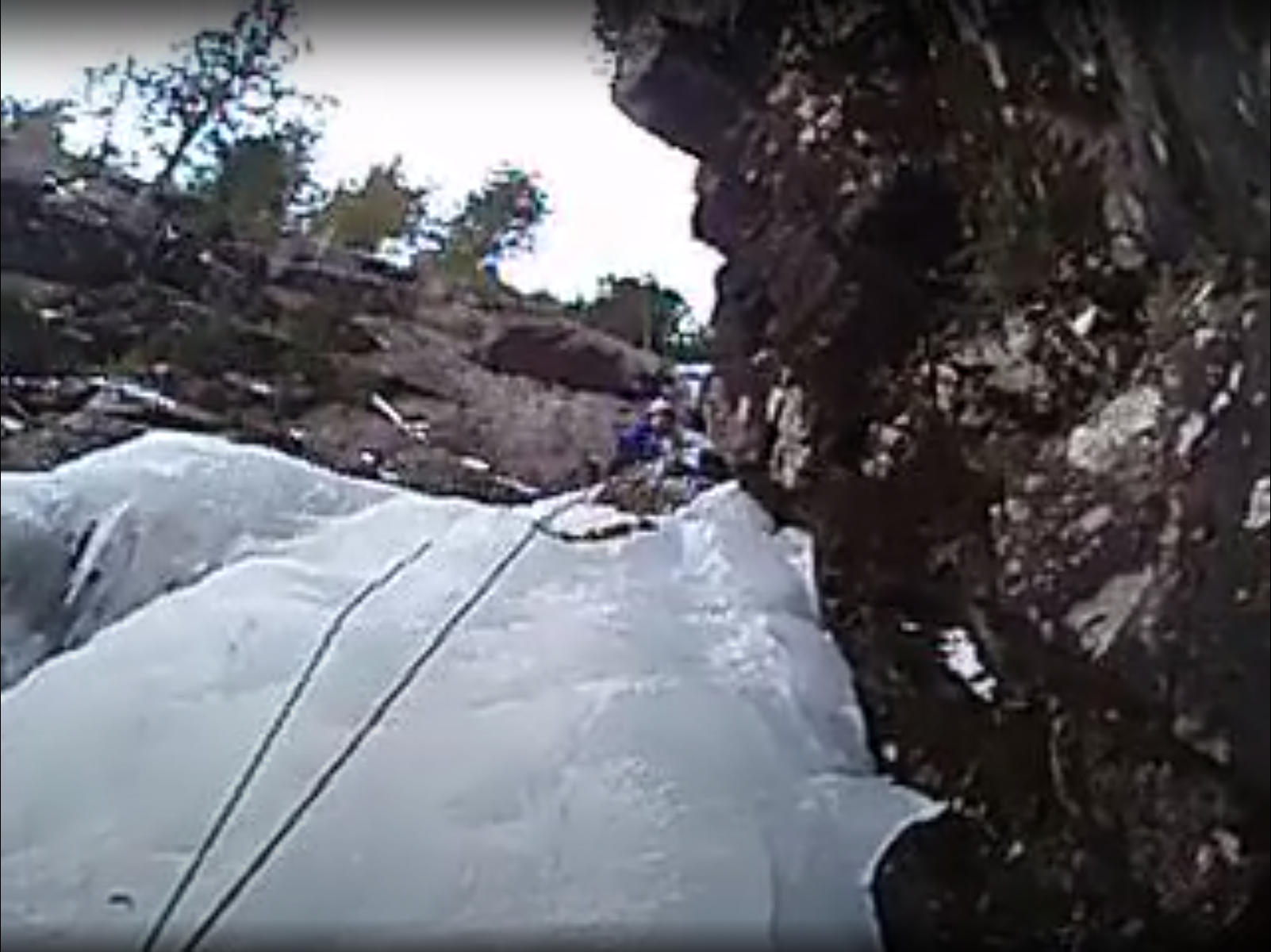} & \includegraphics{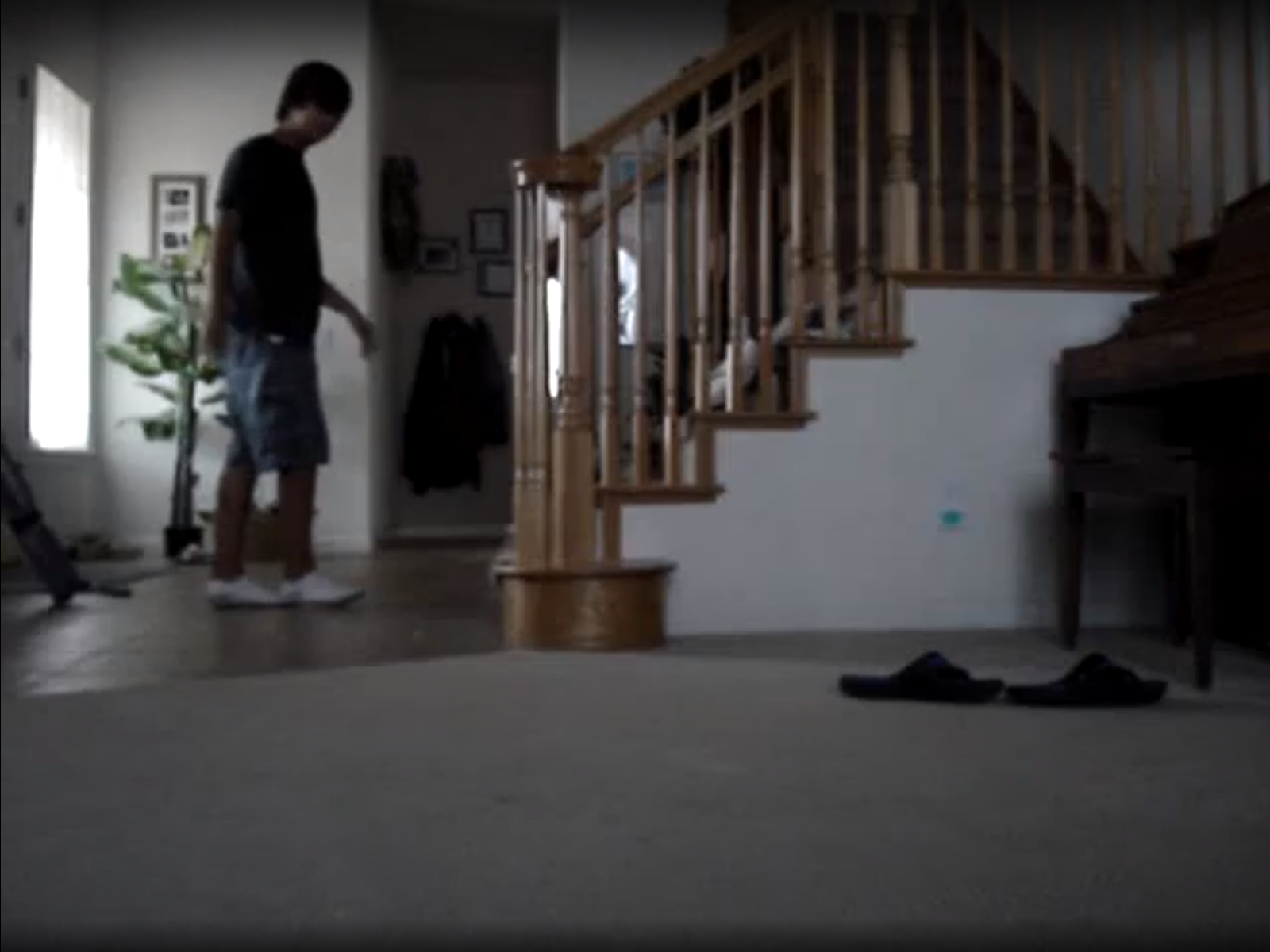} & \includegraphics{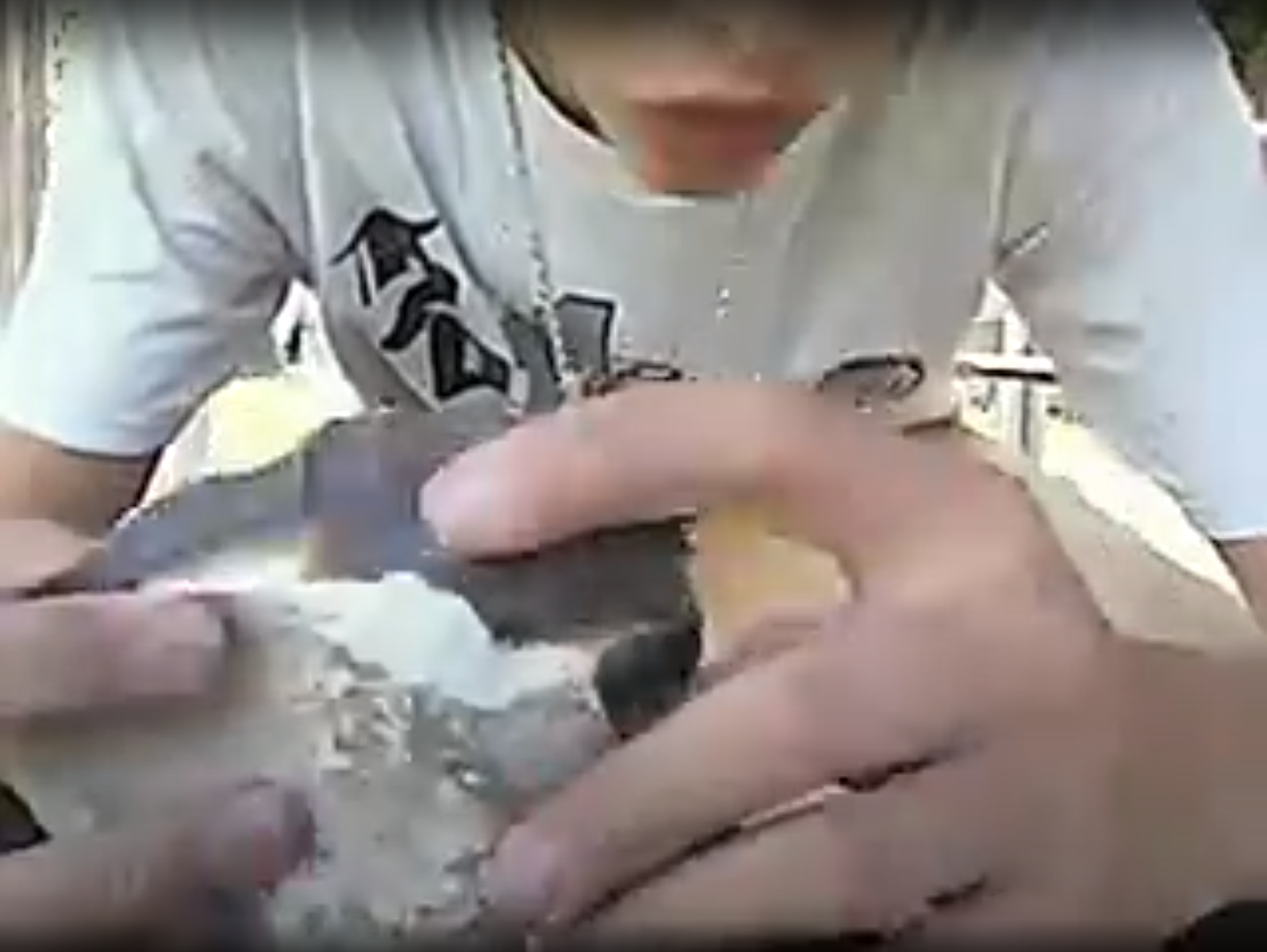} \\
        & {A person at the top of \_\_\_\_\_ with ropes hanging down.} & {A guy is by the stairs in \_\_\_\_\_ doing the moonwalk in socks.} & {A man is showing and describing a rock sample to \_\_\_\_\_.} \\
        \midrule
        correct answers & {adirondacks, cliff, climb, frozen waterfall, gully, hill, ice, icy cliff, ledge, \textbf{mountain}, ravine, slope, snow} & {building, doors, entryway, foyer, his home, his house, home, house, living room, \textbf{room}, shorts, t-shirt} & {audience, \textbf{camera}, consider where its hinge goes, describe how it looks, discuss its hinge, explain his viewers, his audience, his followers, his subscribers, his viewers, people, students, viewer, viewers} \\
        \midrule
        T5 fine-tuned & a tree (0) & a gym (0) & a woman (0) \\
        T5 + I3D & a mountain (100) & a room (100) & a camera (100) \\
    \end{tabular}
\caption{Examples of instances correctly predicted by the best multimodal method but incorrectly predicted by the best text-only method. The F1 score obtained by each answer is shown in parentheses. The correct answers are shown normalized and separated by commas while the model predictions are shown verbatim. From each video, we show a single frame illustrating the key moment.}
\label{tab:qualitative_analysis}
\end{table*}

\end{document}